\documentclass[lettersize,journal]{IEEEtran}
\usepackage{amsmath,amsfonts}
\usepackage{array}
\usepackage[caption=false,font=normalsize,labelfont=sf,textfont=sf]{subfig}
\usepackage{textcomp}
\usepackage{stfloats}
\usepackage{wrapfig}

\usepackage{cite}
\usepackage{url}
\usepackage{microtype}
\usepackage{verbatim}
\usepackage{graphicx}
\usepackage{dirtytalk}
\usepackage[dvipsnames]{xcolor}

\usepackage{indentfirst}
\usepackage[ruled,vlined]{algorithm2e}
\usepackage{amsthm,amssymb}
\usepackage{array}
\newcolumntype{P}[1]{>{\centering\arraybackslash}p{#1}}

\usepackage{makecell}
\usepackage{siunitx, etoolbox, booktabs} 
\usepackage[version=4]{mhchem}
\usepackage{tabularx}
\usepackage{multirow}
\usepackage{bm}
\robustify\bfseries
\usepackage{gensymb}
\usepackage[noend]{algpseudocode}
\usepackage{mathtools}

\def\BibTeX{{\rm B\kern-.05em{\sc i\kern-.025em b}\kern-.08em
    T\kern-.1667em\lower.7ex\hbox{E}\kern-.125emX}}
\usepackage{balance}

\begin{document}

\title{Trajectory Entropy Reinforcement Learning for Robust Robot Motor Skill Learning}

\author{Bang~You, Chenxu~Wang, Wenju~Yang, Di~Guo, Huaping~Liu,~\IEEEmembership{Fellow,~IEEE}
\IEEEcompsocitemizethanks{\IEEEcompsocthanksitem This work was supported by the National Natural Science Fund for Distinguished Young Scholars under Grant 62025304. (\textit{Corresponding author: Huaping~Liu.})}
\IEEEcompsocitemizethanks{\IEEEcompsocthanksitem Bang You is with School
of Information Engineering, Wuhan University of Technology, Wuhan 430070, China. Chenxu Wang and Huaping Liu are with the Department of Computer Science and Technology, Tsinghua University, Beijing 100084, China. Wenju Yang and Di Guo are with School of Artificial Intelligence, Beijing University of Posts and Telecommunications, Beijing 100876, China. E-mail: youbanghust@gmail.com}}

\IEEEspecialpapernotice{
\textnormal{
\textcolor{blue}{This work has been submitted to the IEEE for possible publication.\\
Copyright may be transferred without notice, after which this version may no longer be accessible.}}
\vspace{-2em}
}


\maketitle

\begin{abstract}
Simplicity is a critical inductive bias for designing data-driven controllers, especially when robustness is important. Despite the impressive results of deep reinforcement learning in complex control tasks, it is prone to capturing intricate and spurious correlations between observations and actions, leading to failure under slight perturbations to the environment. To tackle this problem, in this work we introduce a novel inductive bias towards simple policies in reinforcement learning. 
The simplicity inductive bias is introduced by minimizing the entropy of entire action trajectories, corresponding to the number of bits required to describe information in action trajectories after the agent observes state trajectories. Our reinforcement learning agent, Trajectory Entropy Reinforcement Learning, is optimized to minimize the trajectory entropy while maximizing rewards. We show that the trajectory entropy can be effectively estimated by learning a variational parameterized action prediction model, and use the prediction model to construct an information-regularized reward function. Furthermore, we construct a practical algorithm that enables the joint optimization of models, including the policy and the prediction model. Experimental evaluations on several high-dimensional locomotion tasks show that our learned policies produce more cyclical and consistent action trajectories, and achieve superior performance, and robustness to noise and dynamic changes than the state-of-the-art.
\end{abstract}


\section{Introduction}



\IEEEPARstart{D}{eep} reinforcement learning (RL) has emerged as an effective technique to handle hard-to-engineer control tasks across various domains~\cite{chen2022neural, wang2023supervised}, ranging from manufacturing\cite{yasutomi2023visual} to construction~\cite{zhu2023deep}. Without the effort of manually designing state estimators or dynamics models~\cite{you2022integrating, margolis2024rapid}, RL algorithms automatically discover a policy that maps sensory observations into actions. However, these RL approaches tend to learn an intricate mapping from states to actions to achieve satisfying performance. As a result, the learned policies are sensitive to environmental changes and fail when observations or robot dynamics are slightly perturbed, limiting the applications of RL algorithms in robustness-critical scenarios, where achieving robust control is important. 

\begin{figure}
    \centering
    \includegraphics{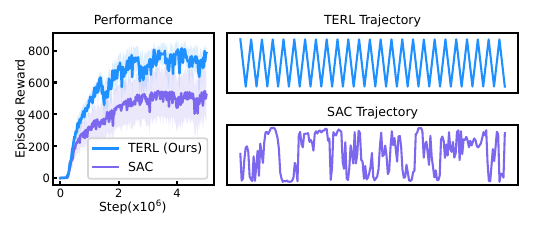}
    \caption{Performance comparison, and action trajectory visualizations of our method (TERL) and its version without the simplicity inductive bias (SAC) on the Humanoid Walk task. Our method learns a policy that produces more periodic and consistent action sequences, improving performance on the Humanoid Walk task.}
    \label{fig:intro_plot}
\end{figure}

In this work, we are interested in learning RL policies that produce simple action sequences to solve control tasks. Consider, for example, the forward walking task of a humanoid robot. Like an animal~\cite{ijspeert2014biorobotics, clark2020predictive}, the robot should move its body in a periodic alternating pattern, rather than erratically (see Figure~\ref{fig:intro_plot}).  Simple policies can avoid unnecessary variations in behaviors, leading to improved robustness to observation noise or dynamic perturbations. 

How can we find RL policies that are simple and effective in solving control tasks at hand? One possibility is to limit the mutual information between policy inputs, states, and its outputs, actions, a metric used to measure model complexity~\cite{leibfried2020mutual, goyal2018infobot, eysenbach2021robust}. The problem is that minimizing the information of states the agent uses for decision-making degrades performance. An alternative is to improve the predictability of future actions based on a history of actions using a prediction model~\cite{saanum2023reinforcement}. However, since these models ignore the use of observed states in predicting actions, they have large prediction errors and cannot effectively measure the predictability of actions.

Entropy can be used to measure the predictability of random variables~\cite{cabrelli1985minimum,watanabe1960information} and has been widely applied in RL~\cite{boularias2011relative, peters2010relative, chen2023entropy, dong2024historical}. Applications of entropy in RL include boosting exploration~\cite{haarnoja2018soft, ziebart2008maximum, igoe2023multi, kim2024accelerating}, and stabilizing training~\cite{peters2010relative}. These works usually focus on entropy maximization at the individual state-action level. However, the use of entropy minimization at the trajectory level, specifically for inducing predictable behaviors, remains largely unexplored.

In this article, we introduce a novel inductive bias towards simple RL policies by minimizing the trajectory entropy, the entropy of entire action trajectories after accessing state sequences.  The trajectory entropy qualifies the number of bits of information required to describe action sequences performed by the policy. Our key intuition is that if the description length of action sequences is short after observing the environment, they are simple and compressible, and therefore robust to environmental perturbations. We use the simplicity inductive bias to guide policy search by minimizing the trajectory entropy, while maximizing rewards. For effectively estimating the trajectory entropy,  we derive a parameterized lower bound of the trajectory entropy using variational inference. Furthermore, we construct an information-
regularized reward function and a practical TERL algorithm based on the lower bound. Policies optimized for maximizing the information-regularized reward function are biased to produce predictable and consistent action sequences, such as cyclical and repetitive gaits. 

We performed extensive experiments on a series of challenging high-dimensional simulated locomotion tasks, such as forward locomotion of Unitree H1~\cite{sferrazza2024humanoidbench}, Humanoid, Cheetah and Walker~\cite{tassa2018deepmind}. Experimental results show that RL policies learned by our method produce more predictable behavior, and perform better and more robustly than state-of-the-art approaches. The key contributions of this work can be summarized as follows:
\begin{itemize}
    \item We introduce a novel trajectory entropy reinforcement learning method, TERL, which minimizes the entropy of action trajectories to induce policies that generate simple behaviors to solve control tasks. 
    \item We derive a parameterized approximation of the trajectory entropy based on variational inference for tractable optimization. 
    \item We propose an information-regularized reward function, which incorporates the preference for predictable action sequences. Furthermore, we propose a practical TERL algorithm based on the reward function. 
    \item A comprehensive evaluations on several high-dimensional locomotion tasks demonstrate that our method improves performance, compressibility of action trajectories, robustness to mass changes, observation noise, and action noise compared to the state-of-art methods~\cite{eysenbach2021robust,saanum2023reinforcement}.
\end{itemize}

Through the above, this work aspires to improve the predictability and robustness of data-driven policies for control, advance the field of RL and facilitate the deployment of robust artificial intelligence in real-world systems, such as legged robots. The remainder of the paper is organized as follows. The related work is discussed in Section~\ref{sec:related_work}. We present the proposed trajectory entropy reinforcement learning method in Section~\ref{sec:method}. Section~\ref{sec: alg} presents the variational approximation of the trajectory entropy, the information-regularized reward, and the TERL algorithm.  Section~\ref{sec:exp_eval} contains details on our implementation of the proposed algorithm and the results of the experimental evaluation. In Section~\ref{sec:conclusion} we draw a conclusion and discuss limitations and future work.

\section{Related Work}
\label{sec:related_work}
The principle of simplicity has received 
significant attention in the machine learning community~\cite{hinton1993autoencoders}. Previous works use this idea to study model pruning~\cite{zhou2010network, wang2021convolutional}, representations learned by neural networks~\cite{tishby2015deep, hu2024survey}, and improve the generalizations of learning-based models~\cite{hinton1993keeping}. Recently, some approaches have applied this principle to the RL setting where the agent can change its action distribution to be simpler. One solution to learning simple policies is to minimize the correlation between states and actions, which is commonly measured by mutual information~\cite{grau2018soft, goyal2018infobot,leibfried2020mutual}. Models representing a simple input-output correlation would be simple. For instance, MIRACLE~\cite{leibfried2020mutual} proposes to limit the mutual information between the current state and the current action, while maximizing rewards. Considering the temporal dimension of decision-making, a recent work, RPC~\cite{eysenbach2021robust}, induces simple policies by minimizing the mutual information between state sequences and their corresponding representation sequences. Minimizing this mutual information allows them to improve the temporal consistency of state representations, implicitly improving the consistency of actions. However, these methods choose actions based on few amount of information about states, leading to poor performance. Our method falls into another class, which facilitates the simplicity of RL policies by explicitly improving the temporal consistency of actions. Typically, a recent pioneering work presented in~\cite{saanum2023reinforcement}, proposes to measure the consistency of actions by the prediction error of an action prediction model that predicts future action based on actions it performed in the past, and uses this prediction error to induce consistent behaviors. However, the action prediction model fails to measure the consistency of actions, since it doesn't use information about environmental observations. Different from these approaches, our method learns simple policies by minimizing the entropy of action trajectories given trajectories of state representations. In Section~\ref{policy visualization}, we will show that our learned policies generate more simpler and consistent behaviors than previous methods.

Previous data-driven approaches learn policies from datasets using inductive biases~\cite{urain2023se,goyal2023rvt,urain2023composable, lee2023robot,yang2024equivact, you2024multimodal}. Huang et al.~\cite{huang2024leveraging} learn the pick-and-place skill from data with the symmetries inductive biases that the actions should be transformed when the observations are rotated or translated, while Char et al.~\cite{char2024pid} improve RL algorithms by introducing the inductive bias performing the summing and differencing operations used in PID controllers. In contrast, we use the simplicity inductive bias to guide the RL agent towards simple solutions (policies).  Some previous approaches introduce entropy-based inductive biases in RL settings and improve the exploration, training stability, and robustness of RL algorithms~\cite{peters2010relative, haarnoja2018soft, yuan2022renyi, wu2020efficient, savas2022entropy, eysenbachmaximum}. However, these objectives usually focus on maximization of an agent’s policy. Different from these approaches, we minimize the entropy of action trajectories, and investigate its effect on inducing simple and robust policies. 

Our method is also related to robust RL~\cite{li2024reinforcement, shi2024distributionally, shi2024robust, zhou2024natural}, which aims to develop RL policies that are resistant to environmental perturbations. While these methods have proposed purpose-designed methods to achieve robustness benefits, we focus on demonstrating that minimizing the trajectory entropy is a simple and effective task-independent solution for improving robustness.

\section{Trajectory Entropy Reinforcement Learning}
\label{sec:method}


In this section, we formulate the problem of learning RL policies generating simple behaviors as a general Markov decision process (MDP). The MDP consists of the state space $s \in \mathcal{S}$, the action space $a \in \mathcal{A}$, the stochastic dynamics model $p(s_{t+1}|s_t,a_t)$, the reward function $r(s,a)$, and the time horizon $T$.
At each time step, the agent observes the current state $s_t$ and chooses its actions $a_t$ based on its stochastic policy $\pi_\theta(a_t|s_t)$ and then receives the reward $r(s_t, a_t)$. The original RL objective is to search for a set of policy parameters $\theta$ that maximize the expected cumulative rewards $\mathbb{E}_{\tau} \left [  \sum_{t=1}^T r_t \right]$ with the agent's trajectory $\tau=(s_1, a_1, s_2, a_2, \dots, a_{T-1}, s_T)$.  As typically not all state information is relevant for choosing the optimal action, we will assume, without loss of generality, that the policy selects the action based on a latent variable $z_t = e_\phi(s_t)$ using an encoder $e_{\phi} : \mathcal{S} \rightarrow \mathbb{R}^d$ with learnable parameters $\phi$.

\begin{figure}
    \centering
    \includegraphics{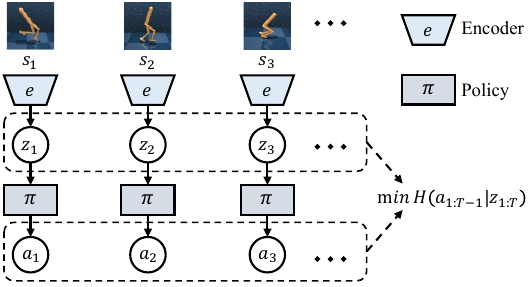}
    \caption{Our TERL agent minimizes the entropy of action trajectories conditioned on trajectories of state representations. The action trajectories are determined by the policy, while the representations are generated by a state encoder. }
    \label{fig:TERL illustration}
\end{figure}

We expect to bias the policy towards generating simple behaviors, achieving robustness benefits. Penalizing the amount of information action sequences contain will keep them simple. Based on this information-theoretic insight, we learn simple policies by maximizing the expected rewards, and meanwhile minimizing the description length of the action sequences after observing state representations sequences:
\begin{equation}
\centering
\begin{aligned}
\max_{\theta, \phi} \quad \mathbb{E}_{\pi_{\theta}, e_{\phi}} \Bigg[ \bigg[  \sum_{t=1}^T   r(s_t, a_t)  \bigg]  - \alpha \mathcal{H}(a_{1:T-1}|z_{1:T}) \Bigg]
\end{aligned}
\label{eq: problem formulation}
\end{equation}
where the hyperparameter $\alpha$ balances behavior complexity and rewards, and the trajectory entropy $\mathcal{H}(a_{1:T-1}|z_{1:T})$ is used to quantify the information cost required to describe the action sequences when the agent knows representation sequences. The trajectory entropy can be defined by
\begin{equation}
\begin{aligned}
\mathcal{H}(a_{1:T-1}|z_{1:T}) = - \mathbb{E}_{p(a_{1:T-1}, z_{1:T})} \Big[ \log p(a_{1:T-1}|z_{1:T}) \Big]
\end{aligned}
\label{eq: entropy definition}
\end{equation}
where the expectation is computed over the joint distribution of action sequences $a_{1:T-1}$ and representation sequences $a_{1:T-1}$. Lower trajectory entropy implies simpler and more predictable action trajectories, and vice versa. We refer to this agent as the Trajectory Entropy Reinforcement Learning (TERL).

We use the conditional entropy $\mathcal{H}(a_{1:T-1}|z_{1:T})$ rather than the entropy of action sequences, because we expect to bias the agent to reduce the complexity of behaviors after knowing the representation sequences. Leveraging information in representations can help predict future actions better and therefore reduce the description length of action sequences. Furthermore, we focus on quantifying the information in action sequences instead of individual actions, allowing us to use expressive predictive models to predict future actions. As shown in Section~\ref{sec: alg}, we will construct a prediction model to predict future action based on the state-action transitions encountered by the agent. 

\section{A Practical TERL Algorithm}
\label{sec: alg}
In this section, we first derive an upper bound on the trajectory entropy for tractable computation. We then modify the rewards by adding the entropy-regularized term. Finally, we construct a practical algorithm for our trajectory entropy reinforcement learning.

\subsection{A Parameterized Upper Bound on trajectory entropy}
Unfortunately, it is non-trivial to directly compute the trajectory entropy, as the conditional distribution $p(a_{1:T-1}|z_{1:T})$ is unknown. For tractable computation, we introduce a variational distribution $q(a_{1:T-1}|z_{1:T})$ to approximate the true conditional distribution $p(a_{1:T-1}|z_{1:T})$ in the definition and obtain an upper bound of the trajectory entropy. Specifically, we first introduce $q(a_{1:T-1}|z_{1:T} )$ into the definition of the trajectory entropy:
\begin{equation}
\begin{aligned}
& \mathcal{H}(a_{1:T-1} | z_{1:T}) = - \mathbb{E}_{p} \bigg[ \log p(a_{1:T-1}|z_{1:T} ) \bigg]\\
&=-\mathbb{E}_{p} \bigg[ \log q(a_{1:T-1}|z_{1:T} ) \bigg] - \mathbb{E}_{p} \bigg[ \log \frac{p(a_{1:T-1}|z_{1:T})}{q(a_{1:T-1}|z_{1:T})} \bigg] \\
&=-\mathbb{E}_{p} \bigg[ \log q(a_{1:T-1}|z_{1:T} ) \bigg] \\
& \quad - \mathbb{D}\textsubscript{KL} \Big(p(a_{1:T-1}|z_{1:T}) \parallel q(a_{1:T-1}|z_{1:T}) \Big)  \\
&\leq - \mathbb{E}_{p} \bigg[ \log q(a_{1:T-1}|z_{1:T} ) \bigg] \\
\end{aligned}
\label{eq: inserting variational distribution}
\end{equation}
where the expectation is computed over the joint distribution $p(z_{1:T},a_{1:T-1})$, and the inequality is introduced by the non-negativity of the Kullback–Leibler (KL) divergence.

We then factor the variational distribution autoregressively:
\begin{equation}
\begin{aligned}
q(a_{1:{T-1}}|z_{1:T}) = \prod_{t=1}^{T-1} q_\psi(a_{t}|z_t, z_{t+1}, a_{t-1})
\end{aligned}
\label{eq: factor autoregressively}
\end{equation}
where $q_\psi(a_{t}|z_t, z_{t+1}, a_{t-1})$ is the variational distribution parameterized by a neural network model with learnable parameters $\psi$. The distribution $q_\psi(a_{t}|z_t, z_{t+1}, a_{t-1})$ can be regarded as the action prediction model, which predicts the current action based on the previous action and the representations of two consecutive states. It at the first time step is degraded to $q_\psi(a_{1}|z_1, z_2)$ that predicts the first action based on the representations at the first and second time steps.

By plugging Eq.~\ref{eq: factor autoregressively} into Eq.~\ref{eq: inserting variational distribution}, we obtain an upper bound on the trajectory entropy for practical optimization:
\begin{equation}
\begin{aligned}
& \mathcal{H}(a_{1:T-1} | z_{1:T}) \leq \mathcal{H}_u(a_{1:T-1} | z_{1:T}) \\
& = - \mathbb{E}_{p} \bigg[ \sum_{t=1}^{T-1} \log q_\psi(a_{t}|z_t, z_{t+1}, a_{t-1})  \bigg]
\end{aligned}
\label{eq: lower bound}
\end{equation}
where the expectation is computed over the joint distribution $p(z_{1:T},a_{1:T-1})$. 

\subsection{Information-regularized Rewards}
By plugging Eq.~\ref{eq: lower bound} into Eq.~\ref{eq: problem formulation}, we obtain the tractable objective to optimize the learning parameters of the policy $\pi_{\theta}$, the encoder $e_{\phi}$, and the action prediction model $q_\psi$ to maximize the rewards and minimize the description lengths of action sequences:

\begin{equation}
\begin{aligned}
\max_{\theta, \phi, \psi} \quad \mathbb{E}_{\pi_{\theta}, e_{\phi}} \Bigg[\sum_{t=1}^{T-1} \Big[ & r(s_t, a_t) + \alpha \log q_\psi(a_{t}|z_t, z_{t+1}, a_{t-1}) \Big] \\
& + r(s_T, a_T) 
\Bigg].
\end{aligned}
\end{equation}

Our trajectory entropy objective modifies the reward function by adding the information cost term, which corresponds to minimizing the amount of information required to describe actions. The policy now is to maximize the information-regularized reward:
\begin{equation}
\begin{aligned}
& r^* (s_t, a_t) = r(s_t, a_t) + \alpha \Big(\log q_\psi(a_{t}|z_t, z_{t+1}, a_{t-1}) \Big).
\end{aligned}
\label{eq:information-regularized reward}
\end{equation}

The modified reward incentivizes the agent to perform actions, which can be predicted well using the previous action and the representations of two consecutive states. Actions can be easily predicted by the prediction model if they contain cyclical and consistent structures.

\subsection{Algorithm Implementations}

For the practical implementation, we transfer the finite-horizon problem into the infinite one by introducing the discount factor $\gamma$ and setting the horizon $T$ to infinite, and obtain the final objective:
\begin{equation}
\begin{aligned}
\max_{\theta, \phi, \psi} \quad &  \mathbb{E}_{\pi_\phi, e_\theta} \Bigg[\sum_{t=1}^\infty \gamma^t r^*(s_t, a_t)
\Bigg]\\
\end{aligned}
\label{eq: final obj}
\end{equation}

\begin{figure}
    \centering
    \includegraphics{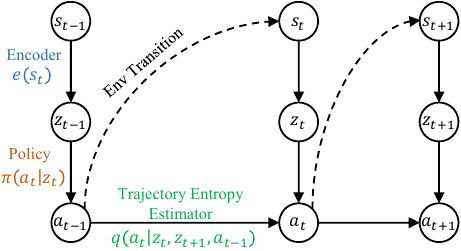}
    \caption{We use one objective to jointly optimize our policy, the encoder, and the lower bound of the trajectory entropy.  }
    \label{fig:actor objective}
\end{figure}

Our trajectory entropy reinforcement learning objective in Eq.~\ref{eq: final obj} can be optimized using existing RL methods. In our implementation, we implement TERL as an extension of the soft actor-critic algorithm (SAC)~\cite{haarnoja2018soft}, which proceeds by alternating between policy improvement and policy evaluation. For the policy evaluation of our algorithm, we do not make any changes to SAC, except for replacing the original reward function $r(s_t, a_t)$ with the information-augmented reward $r^*(s_t, a_t)$. The Q function with parameters $\upsilon$ is optimized by minimizing the temporal difference loss:
\begin{equation}
\begin{aligned}
L\left(\upsilon \right)=\mathbb{E}_{\mathcal{D}, \pi_{\theta}, e_{\phi}} \left[\big(Q_{\upsilon}(s_t, a_t)- y(s_t, a_t)\big)^{2}\right]
\end{aligned}
\label{eq: Q loss}
\end{equation}
where the target  is given by
\begin{equation}
\begin{aligned}
y(s_t, a_t) &=   r^* (s_t, a_t) + \gamma(1-d) \big[Q_\upsilon(s_{t+1}, a_{t+1}) \\
& - \beta \log\big(\pi_{\theta}(a_{t+1}|z_{t+1})\big) \big]\\
\end{aligned}
\end{equation}
with termination flag $d$. The next action $a_{t+1}$ is sampled from the policy and the current state-action pair is sampled from replay buffer $\mathcal{D}$. $\beta$ is the coefficient of the policy entropy term in SAC.

For our policy improvement, following LZ-SAC~\cite{saanum2023reinforcement}, we express the Q function as the immediate information-augmented reward and the Q function of the next timestep, and obtain the objective to jointly optimize the policy, the encoder, and the action prediction model: 

\begin{equation}
\begin{aligned}
& \max_{\theta, \phi, \psi}  \quad   \mathbb{E}_{\mathcal{D}, \pi_{\theta}, e_{\phi}}  \bigg[ r(s_t, a_t) + \alpha \log q_\psi(a_{t}|z_t, z_{t+1}, a_{t-1})  \\
& - \beta \log \pi_{\theta}(a_t|z_t) + \gamma \Big( Q_\upsilon(s_{t+1}, a_{t+1}) - \beta \log \pi_{\theta}(a_{t+1}|z_{t+1}) \Big) \bigg]
\end{aligned}
\label{eq:actor loss}
\end{equation}
where the state-action pair $[s_t, a_t, s_{t+1}]$ is sampled from replay buffer $\mathcal{D}$, and the next action $a_{t+1}$ is sampled from the policy, and the representations, $z_t$ and $z_{t+1}$, are generated by the encoder. We illustrate the objective in Eq.\ref{eq:actor loss} in Figure.~\ref{fig:actor objective}.

The procedure of our TERL algorithm is shown in Algorithm~\ref{alg:alg_procedure}. The algorithm proceeds by alternating between collecting new experience from the environment, and updating the parameters of the Q-function, the policy, the encoder, and the auxiliary prediction model.

\begin{algorithm}[ht]
\caption{TERL}
\label{alg:alg_procedure}
\SetAlgoLined
 \textbf{Initialize:} policy $\pi_\theta$, encoder $e_{\phi}$, action prediction model $q_\psi$, Q function $Q_\upsilon$, replay buffer $\mathcal{D}$,  coefficients $\alpha, \beta$, batch size $B$, learning rate $\rho$\\
 \For{each training step}{
  collect experience $(s_t, a_t, r_t, s_{t+1})$ and add it to replay buffer\\
  \For{each gradient step}{
   Sample a minibatch of transitions from replay buffer: $(s_t, a_t, r_t, s_{t+1}, a_{t-1}) \sim \mathcal{D}$ \\ 
   Compute upper bound: $u \leftarrow  \mathbb{E}\Big[\log q_\psi(a_{t}|z_t, z_{t+1}, a_{t-1})\Big]$ \Comment{Eq.~\ref{eq: lower bound}}\\
   Compute information-regularized reward:  $r^* \leftarrow r + \alpha u$ \Comment{Eq.~\ref{eq:information-regularized reward}}\\
   Update Q function: $\upsilon \leftarrow \upsilon - \rho \hat{\nabla}_{\upsilon} \mathcal{L}(\upsilon)$ \Comment{Eq.~\ref{eq: Q loss}}\\
   Update policy, encoder, and prediction model: $\{\theta, \phi, \psi\}  \leftarrow \{\theta, \phi, \psi\} - \rho \hat{\nabla}_{\{\theta, \phi, \psi\}} \mathcal{L}({\theta, \phi, \psi})$ 
   \quad \quad \Comment{Eq.~\ref{eq:actor loss}}\\
  }
 }
\end{algorithm}

\section{Experimental Evaluation}
\label{sec:exp_eval}
In this section, we perform extensive experiments to evaluate our approach. Our experiments aim to investigate TERL from the following perspectives: (1) we investigate whether minimizing the trajectory entropy helps policy learning (Sec.~\ref{performance}). (2) we study how the learning performance of TERL compares to the state-of-the-art approaches (LZ-SAC and RPC) (Sec.~\ref{performance}). (3) we examine the empirical properties of simple policies learned by our method, such as their zero-shot robustness (Sec.~\ref{robustness}) and ability to produce simple behaviors to solve control tasks (Sec.~\ref{policy visualization}). (4) we investigate the effect of the hyperparameters of TERL on inducing simple behaviors, such as the hyperparameter $\alpha$ balancing rewards and trajectory entropy (Sec.~\ref{ablation}). 

\subsection{Experimental Setup}
\begin{figure}
    \centering
    \includegraphics{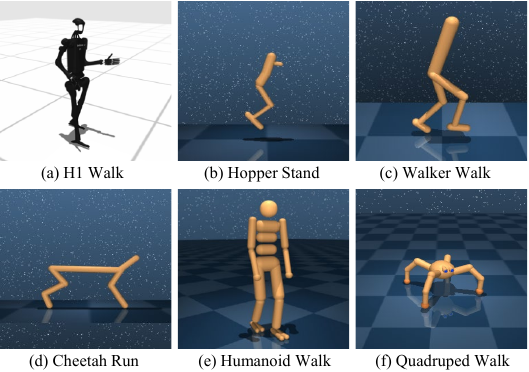}
    \caption{We evaluate our method and previous methods on six high-dimensional locomotion tasks: H1 Walk, Hopper Stand, Walker Walk, Cheetah Run, Humanoid Walk, and Quadruped Walk. }
    \label{fig:tasks}
\end{figure}

We evaluate our method on several challenging high-dimensional locomotion tasks from DMC Suite~\cite{tassa2018deepmind} following LZ-SAC~\cite{saanum2023reinforcement}, which is a commonly used control benchmark, and HumanoidBench~\cite{sferrazza2024humanoidbench}, a recent benchmark focusing on complex control tasks for a Unitree H1 humanoid robot. Specifically, we consider six locomotion tasks, Hopper Stand, Walker Walk, Cheetah Run, Humanoid Walk, Quadruped Walk from DMC Suite, and H1 Walk from HumanoidBench (see Figure~\ref{fig:tasks}). Each task provides challenges, including high-dimensional action space, and contacts with ground. More detailed descriptions of these tasks are available in~\cite{tassa2018deepmind} and~\cite{sferrazza2024humanoidbench}.

We compare our method with LZ-SAC~\cite{saanum2023reinforcement}, which learns a simple policy by improving compressibility of action sequences and achieves leading performance on DMC tasks, RPC~\cite{eysenbach2021robust}, which induces simple behaviors by minimizing the mutual information between a sequence of states and a sequence of their representations, and SAC~\cite{haarnoja2018soft}, which is a competitive RL algorithm that achieves good performance on control tasks.

\begin{table}[hbt!]
    \centering
    \caption{Description of the used experimental tasks.}
    \label{table:task description}
    \begin{tabular}{c|c | c}
        \Xhline{2\arrayrulewidth}
        Tasks & State Dimension & Action Dimension\\
        \hline
        H1 Walk & 51 & 19\\
        Hopper Stand & 15 & 4\\
        Walker Walk & 24  & 6\\
    Cheetah Run & 17 & 6\\
    Humanoid Walk & 67 & 21\\
    Quadruped Walk & 78 & 12\\
        \Xhline{2\arrayrulewidth}
    \end{tabular}
\end{table}

\begin{table}[hbt!]
    \centering
    \caption{Hyperparameters used for TERL.}
    \label{table:TERL hyper}
    \begin{tabular}{c|P{30mm}}
        \Xhline{2\arrayrulewidth}
        Parameter & Value\\
        \hline
        Replay buffer capacity & 100 0000\\
        Optimizer & Adam\\
        Actor learning rate & $10^{-4}$ \\
        Actor update frequency & 2\\
        Actor log stddev bounds & [-10 2]\\
        Critic Learning rate & $10^{-4}$ \\
        Critic  Q-function EMA & 0.01\\
        Critic target update freq & 2\\
        Temperature learning rate & $10^{-4}$ \\
        Initial steps & 5000\\
        Discount & 0.99\\
        Initial temperature & 0.1\\
        Learning rate for TERL & $10^{-4}$ \\
        Batch size & 256\\
        Coefficient $\alpha$ for Walker Walk & $10^{-4}$\\
        Coefficient $\alpha$ for other tasks & $10^{-5}$\\
        \Xhline{2\arrayrulewidth}
    \end{tabular}
\end{table}

We implement our method based on the commonly open-sourced Pytorch implementations of the SAC algorithm~\cite{yarats2019improving}. While the original implementation of LZ-SAC is built on top of the same SAC implementation, the official implementation of RPC uses the SAC algorithm from TF-Agents. To ensure a fair and reliable comparison, we compare our method to RPC implemented by its original implementation (referred to as RPC-Orig) and to our RPC implementation built on top of the same SAC codebase with LZ-SAC and TERL (referred as RPC). 

We parameterize the encoder $e_\phi$ and the action prediction model $q_\psi$ both using a 3-layer neural network with the ReLU activation function and a hidden dimension of 256. The output dimension is set to 30 for the encoder. The output of the action prediction is divided into the mean and the standard deviation of a diagonal Gaussian distribution used for predicting actions. We determine the hyperparameter $\alpha$ by performing hyperparameter tuning. Specifically, we define a set of the $\alpha$ values, $[10^{-5}, 10^{-4}]$, and perform a grid search over it on all tasks. Based on the empirical results, we set it to $10^{-5}$ for all tasks, except for the Walker Walk task where $\alpha$ is set to $10^{-4}$.  We use the default SAC parameters from the implementations~\cite{yarats2019improving} for each method, unless specified otherwise. All learnable parameters of our model are updated using the Adam optimizer with a learning rate of $10^{-4}$. We provide an overview of all hyperparameters in Table~\ref{table:TERL hyper}. 

 For each method, the agent performs one gradient update per environment step to ensure a fair comparison. For a more reliable comparison, we run each algorithm with 20 independent seeds for each task. We train all agents for $10^{6}$ steps and evaluate them every 20000 steps.  

\subsection{Performance}
\label{performance}

\begin{table*}[ht!]
\caption{Scores (means over 20 seeds with 90\% confidence interval) achieved by our method and baselines on 6 locomotion tasks. TERL achieves higher average scores than previous methods on the majority of tasks.}
\begin{center}
\sisetup{%
            table-align-uncertainty=true,
            separate-uncertainty=true,
            detect-weight=true,
            detect-inline-weight=math
        }
{\begin{tabular}{c | c c c c c c c}
    \Xhline{2\arrayrulewidth}
    Scores & TERL (Ours) & RPC  & RPC-Orig & LZ-SAC & SAC\\
    \hline
    H1 Walk & \bfseries 137$\pm$\bfseries 19 & 26$\pm$5 & 34$\pm$12 & 5$\pm$1 &  85$\pm$17\\
    Hopper Stand & \bfseries 920$\pm$\bfseries 19 & 568$\pm$96  & 476$\pm$101 & 593$\pm$88 & 683$\pm$114\\
    Walker Walk & \bfseries 972$\pm$\bfseries 2  & 940$\pm$21 & 951$\pm$2 & 939$\pm$26 & 962$\pm$7\\
    Cheetah Run & \bfseries 842$\pm$\bfseries 31  & \bfseries 772$\pm$ \bfseries 57 & 636$\pm$10 & 787$\pm$17 &  \bfseries 811$\pm$ \bfseries 36\\
    Humanoid Walk & \bfseries 479$\pm$\bfseries 52 &  53$\pm$48 & 2$\pm$1 &  1$\pm$0 &   317$\pm$  93\\
    Quadruped Walk & \bfseries 918$\pm$\bfseries 26 &  \bfseries 842$\pm$\bfseries 77 &  598$\pm$108 & 595$\pm$110 &  738$\pm$93\\
    \Xhline{2\arrayrulewidth}
\end{tabular}}
\end{center}
\label{table:performance}
\end{table*}


\begin{table}[ht!]
\caption{Performance drops in percentage caused by mass changes. Bold font indicates the highest performance percentage among all methods.}
\begin{center}
{\begin{tabular}{c | c c c c c c c}
    \Xhline{2\arrayrulewidth}
    Mass Scale & TERL (Ours) & RPC  & LZ-SAC & SAC\\
    \hline
    s=0.50 & \bfseries 75.5 & 73.7 & 66.0 & 69.7\\
    s=0.75 & \bfseries 93.5 & 91.0  & 89.7 & 89.6\\
    s=1.25 & \bfseries 96.7  & 95.7 & 93.7 & 94.4\\
    s=1.50 & \bfseries 80.3  & 78.7 & 78.6 & 73.2\\
    \Xhline{2\arrayrulewidth}
\end{tabular}}
\end{center}
\label{table: mass performance drops}
\end{table}

\begin{table}[ht!]
\caption{Performance drops in percentage caused by action noise. Bold font indicates the highest performance percentage among all methods.}
\begin{center}
{\begin{tabular}{c | c c c c c c c}
    \Xhline{2\arrayrulewidth}
    Action Noise Strength & TERL (Ours) & RPC  & LZ-SAC & SAC\\
    \hline
    $\sigma$=0.05 & \bfseries 99.1 & 98.9  & 98.7 & 98.8\\
    $\sigma$=0.10 & \bfseries 98.6 & 97.9  & 94.1 & 97.6\\
    $\sigma$=0.15 & \bfseries 97.0  & 96.3 & 87.0 & 95.6\\
    $\sigma$=0.20 & \bfseries 95.0  & 94.8 & 79.9 & 93.8\\
    $\sigma$=0.25 & \bfseries 92.7 & 92.6 & 74.2 & 92.1\\
    $\sigma$=0.30 & 90.1 & \bfseries 90.7  & 68.4 & 89.1\\
    \Xhline{2\arrayrulewidth}
\end{tabular}}
\end{center}
\label{table: action performance drops}
\end{table}

\begin{table}[ht!]
\caption{Performance drops in percentage caused by observation noise. Bold font indicates the highest performance percentage among all methods.}
\begin{center}
{\begin{tabular}{c | c c c c c c c}
    \Xhline{2\arrayrulewidth}
    Obs Noise Strength & TERL (Ours) & RPC  & LZ-SAC & SAC\\
    \hline
    $\sigma$=0.02 & \bfseries 97.6 & 97.0 & 95.7 & 95.5\\
    $\sigma$=0.04 & \bfseries 92.9 & 91.9  & 83.9 & 90.7\\
    $\sigma$=0.06 & \bfseries 87.6 & 87.0  & 78.1 & 86.1\\
    $\sigma$=0.08 & \bfseries 83.5  & 83.4 & 74.
    4 & 81.6\\
    $\sigma$=0.10 & 78.8  & \bfseries 79.9 & 73.1 & 77.7\\
    \Xhline{2\arrayrulewidth}
\end{tabular}}
\end{center}
\label{table: obs performance drops}
\end{table}

Table~\ref{table:performance} provides a comparison of TERL to LZ-SAC, RPC and SAC on 6 locomotion tasks. Scores in the Table are averaged over 20 independent runs, with the error bar representing 90\% confidence interval of the standard error of the mean. TERL outperforms the state-of-the-art methods on the majority of tasks in terms of performance, showcasing its ability to better handle challenging locomotion tasks. On the high-dimensional H1 Walk task ($\mathcal{S} \in \mathbb{R}^{51}$, $\mathcal{A} \in \mathbb{R}^{19}$), for instance, our method achieves an average reward of 137, significantly higher than 34 achieved by RPC, and 5 achieved by LZ-SAC. Similar results are observed in the Humanoid Walk task ($\mathcal{S} \in \mathbb{R}^{67}$, $\mathcal{A} \in \mathbb{R}^{21}$), where our method achieves an average reward of 479, while LZ-SAC and RPC obtain rewards of 1 and 53, respectively. Predictable policies can avoid unnecessary variations in actions and may improve the rewards on these control tasks. Hence, we conjecture that the performance improvement achieved by TERL is caused by the improved predictability of policies by limiting the trajectory entropy. We will investigate the predictability of policies in Section~\ref{policy visualization}. Notably, our method outperforms SAC by a large margin in five of six tasks. SAC can be regarded as a special case of our TERL algorithm where no simplicity inductive bias is used ($\alpha = 0$). The performance improvement achieved by our method than SAC indicates that minimizing trajectory entropy helps to learn good policies on locomotion tasks. 

\begin{figure*}
    \centering
    \includegraphics{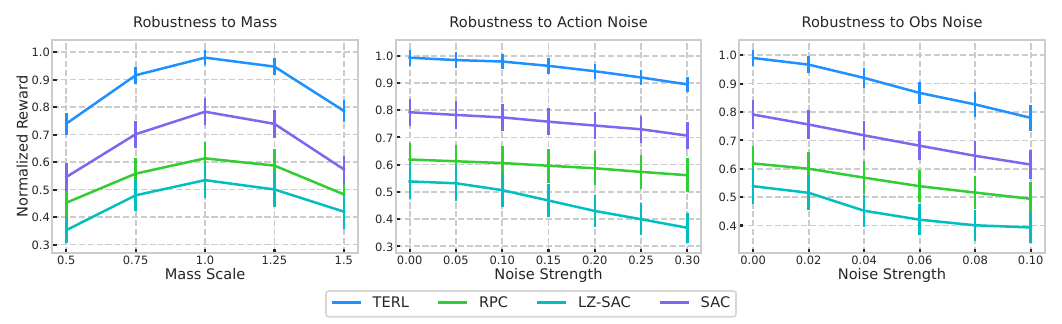}
    \caption{Zero-shot robustness to changes in body gravity (left), action noise (middle), and observation noise (right) on 6 locomotion control tasks. This plot shows the normalized mean rewards averaged over 20 independent runs and 6 tasks, with error bar representing 90\% confidence interval. To make comparisons across tasks, we normalize rewards by the reward achieved by the best method on each task. For each run, we collect 30 evaluation trajectories. TERL achieves better aggregated performance than all baselines when environments are disturbed by action noise, mass changes, and small observation noise.}
    \label{fig:robustness_results}
\end{figure*}

\subsection{Robustness}
\label{robustness}
Our policies are biased to produce behaviors that are easily predictable and have fewer variations, and therefore we expect they are more robust to unseen environment perturbations. In this section, we evaluate the learned policies by our method and baselines in robustness to mass changes, action noise, and observation noise. 

\subsubsection{Robustness to Mass Change}
We first investigate the robustness to deviations between the dynamics encountered during testing and the dynamics used for training. We introduce the dynamic mismatch by scaling the gravity of the robot body using a fixed factor during evaluation. Using the same six tasks, we evaluate our method and baselines on a series of mass scaling factors, $s \in [0.5, 0.75, 1.25, 1.5]$. To compare the robustness across all six tasks, we normalized the scores by the score achieved by the best method on each task.  We obtain the aggregated result on all six tasks by computing the normalized average reward over 20 seeds and 6 tasks. Figure~\ref{fig:robustness_results} (left) shows the aggregated robustness results for 6 tasks with different mass scaling factors. Policies learned by TERL achieve the highest rewards among all methods when the robot's body mass is changed. For example, when increasing the robot mass with a scale of 1.5, TERL obtains an average reward of around 0.8, while other approaches achieve rewards lower than 0.6. When decreasing the mass with a scale of 0.5, the average reward for TERL is 0.74, 1.6 times higher than the reward of 0.45 for RPC and 2.1 times higher than the reward of 0.35 for LZ-SAC. 

We also compare the performance drops in percentage caused by mass changes in Table~\ref{table: mass performance drops}. For each method, we compute the percentages by the achieved average rewards in the presence of mass perturbations divided by the average rewards without mass perturbations. We observed 
that rewards achieved by our method decrease more slowly than RPC, LZ-SAC, and SAC for all 4 different mass scaling factors. Based on the results in Figure~\ref{fig:robustness_results} and Table~\ref{table: mass performance drops}, we conclude that TERL is more robust than the previous approaches when the robot mass is changed.

\subsubsection{Robustness to Action Noise}
An agent's action may be changed during execution due to system inaccuracies. We study the robustness to action noise by adding Gaussian noise into the actions during evaluation, where noise is sampled from a Gaussian distribution $\mathcal{N}(0, \mathrm{diag}(\sigma^2))$ with noise strength $\sigma$.  For each task, we test all methods on six different noise strengths, $\sigma \in [0.05, 0.1, 0.15, 0.2,0.25, 0.3]$. We present the aggregated normalized result over 20 seeds and 6 locomotion tasks in  Figure~\ref{fig:robustness_results} (middle). Overall, our method obtains higher mean rewards than baselines in the presence of action noise. When adding noise with a strength of 0.3, for instance, TERL achieves a mean reward of 0.89, 2.4 times higher than the reward of 0.37 achieved by LZ-SAC, and 1.6 times higher than the reward of 0.56 achieved by RPC. 

Table~\ref{table: action performance drops} compares the performance drops in percentage caused by action noises for 6 different noise strengths. Each percentage in Table~\ref{table: action performance drops} is computed by the achieved average rewards in the presence of action noises divided by the average rewards without action noises. Experimental results show that the performance drops of TERL are slower than baselines in the presence of small action noises. This indicates that TERL outperforms previous approaches in the robustness to action perturbations.

\subsubsection{Robustness to Observation Noise}
We also investigate the effect of observation noise on performance. We add Gaussian noise to the states during evaluation, where noise is sampled from a Gaussian distribution. Using the same six tasks, we evaluate our method and baselines on five different noise strengths, $\sigma \in [0.02, 0.04, 0.06, 0.08, 0.10]$, and present the aggregated normalized results in Figure~\ref{fig:robustness_results} (right). TERL outperforms other methods while states are disturbed by Gaussian noise. For example, TERL obtains a reward of 0.92, significantly higher than a reward of 0.57 achieved by RPC and a reward of 0.72 achieved by SAC, when adding Gaussian noise with a strength of 0.04. 

Table~\ref{table: obs performance drops} compares the performance drops in percentage caused by observation noises for 5 different noise strengths. We compute each percentage in Table~\ref{table: action performance drops} by dividing the achieved average rewards in the presence of observation noises by the average rewards without observation noises. TERL achieves smaller performance drops than baselines in the presence of small observation noises. We conclude that TERL is more robust than baselines in the presence of observation perturbations.

\begin{table}[b!]
\caption{We evaluate TERL with different hyperparater $\alpha$ on the Walker Walk task. The compressibility of behaviors can be improved while the hyperparameter $\alpha$ increases.}
    \centering
    \begin{tabular}{c|c c c}
    \Xhline{2\arrayrulewidth}
    Hyperparameter  & $\alpha=0.0001$ & $\alpha=0.001$ & $\alpha=0.01$ \\
    \hline
      Normalized Bytes   & 1.00 $\pm$ 0.02 & 0.98 $\pm$ 0.01  & \bfseries  0.50 $\pm$ \bfseries 0.03 \\
    \Xhline{2\arrayrulewidth}
    \end{tabular}
\label{tab:hyperparameter_compression}
\end{table}

We attribute the achieved robustness improvements to the fact that TERL produces the most compressive action trajectories (see Section~\ref{policy visualization}), since compressive behaviors contain fewer variations and thus are less sensitive to environmental perturbations. 

\begin{figure*}
    \centering
    \includegraphics{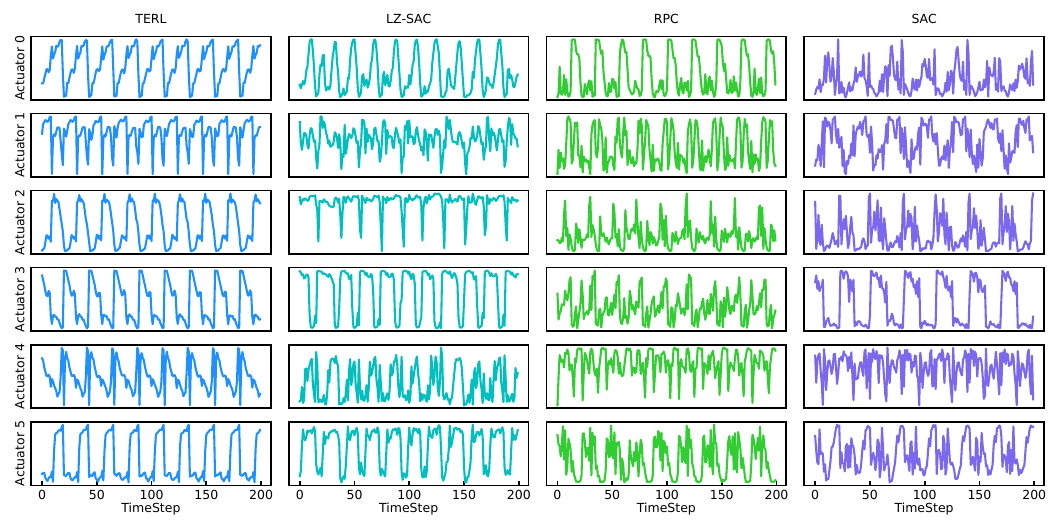}
    \caption{We visualize action sequences generated by our method and baselines on the Walker Walk task. Behaviors produced by our agent show more periodic and predictable patterns than baselines.}
    \label{fig:traj visualization}
\end{figure*}

\begin{figure*}
    \centering
    \includegraphics{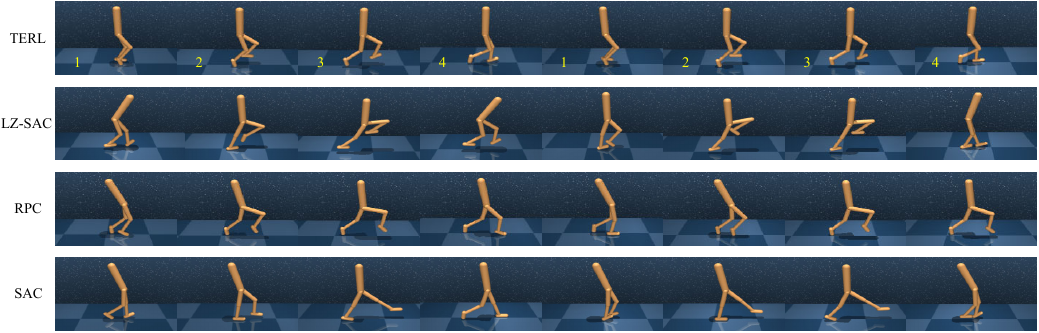}
    \caption{Visualizations of the gaits produced by our method and baselines on the Walker Walk task. Gaits produced by our method show more cyclical and predictable patterns than previous approaches.}
    \label{fig:configuration visualization}
\end{figure*}

\begin{figure}
    \centering
    \includegraphics{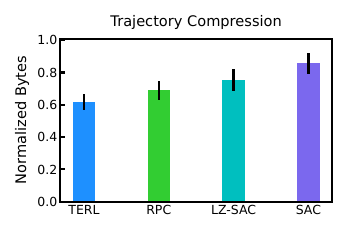}
    \caption{We compress trajectories produced by TERL and baselines on all DMC tasks by a lossless compression algorithm, bzip2. The plot shows normalized averaged file size in bytes over 20 runs and 5 tasks, with error bar representing 90\% confidence interval. Each run includes 30 collected trajectories. TERL achieves the smallest average size among all methods. }
    \label{fig:traj compression} 
\end{figure}

\subsection{Policy Visualization}
\label{policy visualization}
To inspect the simplicity of behaviors, we first visualize the trajectories produced by learned policies. Figure~\ref{fig:traj visualization} presents action trajectories produced by our method, LZ-SAC, RPC, and SAC on the Walker Walk task. TERL produces the most consistent and simplest trajectory over time, featuring a high degree of periodicity. In contrast, the trajectories produced by LZSAC and RPC show fewer cyclical patterns than TERL, while the trajectory generated by SAC exhibits numerous erratic changes and fluctuations. This verifies our hypothesis that minimizing the description length of action sequences can improve the simplicity of policies.

We further examine the locomotion gaits produced by our method and baselines in Figure~\ref{fig:configuration visualization}. Our TERL policy produces a highly repetitive and consistent gait cycle, which can be divided into 4 phases (see sequences of robot configurations labeled 1 to 4 in the top row), namely, putting two legs together, lifting the left leg, moving forward, and moving the right leg. The movements produced by LZ-SAC (second row) and RPC (third row) do not follow a clear, repetitive pattern. Lastly, the SAC policy produces the most erratic gaits. The improved repetition in the walking movements achieved by our method verifies the effectiveness of our trajectory entropy objective in inducing simple policies.

Furthermore, motivated by~\cite{saanum2023reinforcement}, we use lossless compression algorithms to \emph{quantify} the compressibility of trajectories. Trajectories can be effectively compressed when they contain repetitive or periodic structures. For each task, we collect state-action trajectories using learned policies, round them to one decimal place, save them as .npy file, and then compress them using bzip2, a common lossless compression algorithm. Figure~\ref{fig:traj compression} shows the normalized average file sizes in bytes over 20 runs and 5 DMC tasks for all methods. The normalized file sizes are achieved by scaling the file sizes by the largest size among all methods for each task. We do not consider the H1 Walk task from HumanoidBench in this experiment, as the trajectory horizon is not fixed for this task. Experimental results in Figure~\ref{fig:traj compression} show that trajectories produced by TERL are more easily compressed than other methods, indicating that our learned policies generate trajectories that show more cyclical and repetitive patterns to solve tasks. 

\subsection{Hyperparameter Ablation}
\label{ablation}

In this section, we perform an ablation study to investigate the effect of the hyperparameter $\alpha$ on inducing compressive trajectories. The hyperparameter $\alpha$ controls the trade-off between maximizing rewards and minimizing the trajectory entropy, and increasing the value of $\alpha$ will bias the agent to decrease the trajectory entropy. We evaluate our method with three different $\alpha$ in terms of the compressibility of trajectories on the Walker Walk task. 

Table~\ref{tab:hyperparameter_compression} shows the normalized average file size over 20 runs, with error bars representing 90\% confidence interval. Increasing the value of $\alpha$ from $0.0001$ and $0.001$ to $0.01$ dramatically decreases the size of trajectories. This suggests that minimizing the upper bound of the trajectory entropy helps induce compressible behaviors.

\section{Conclusion and Discussion}
\label{sec:conclusion}

We presented Trajectory Entropy Reinforcement Learning, TERL, which uses the objective of trajectory entropy as a regularizer to induce predictable and robust RL policies. We derive an upper bound of the trajectory entropy for tractable optimization and use it to construct a practical TERL algorithm. Our learned policies produce behaviors containing highly cyclical and consistent patterns. Experimental results on several simulated locomotion tasks show that our method outperforms leading baselines in terms of performance, compressibility of trajectories, and robustness to mass changes, action noise, and observation noise.

While achieving promising performance,  TERL shares a limitation common to methods with predictable policies: the simplicity inductive bias can for some tasks harm the performance of the learned policy, namely, when the task inherently demands complex and unpredictable behaviors, e.g. playing poker.  Consistent and predictable behaviors, however, are highly desirable in some RL applications, such as deploying reinforcement learning agents to interact with humans. Applying our TERL algorithm to solve human-machine interaction tasks is a direction for our future work. Additionally, we rely on grid search to determine the value of the hyperparameter $\alpha$, leading to high computational costs. In our future work, we will explore leveraging task-specific information, such as the dimension of the state space, to automatically choose the coefficient of our trajectory entropy objective. Moreover, developing additional upper bounds of our trajectory entropy objective to further improve performance presents another avenue for future research.


\bibliography{references}


%

\ifCLASSOPTIONcaptionsoff
  \newpage
\fi



\end{document}